\documentclass{article}

\usepackage[preprint]{neurips_2023}

\usepackage[utf8]{inputenc} 
\usepackage[T1]{fontenc}    
\usepackage{hyperref}       
\usepackage{url}            
\usepackage{booktabs}       
\usepackage{amsfonts}       
\usepackage{nicefrac}       
\usepackage{microtype}      
\usepackage{xcolor}         

\usepackage{microtype}
\usepackage{graphicx}
\usepackage{subfigure}
\usepackage{booktabs} 

\usepackage{diagbox}

\usepackage{bm}

\usepackage{amsmath}
\usepackage{amssymb}
\usepackage{mathtools}
\usepackage{amsthm}

\theoremstyle{plain}

\theoremstyle{definition}

\theoremstyle{remark}

\usepackage[textsize=tiny]{todonotes}

\title{IQL-TD-MPC: Implicit Q-Learning for\\Hierarchical Model Predictive Control}

\author{%
  Rohan Chitnis\thanks{equal contribution} \\
  Meta AI\\
  \texttt{ronuchit@meta.com} \\
  \And
  Yingchen Xu\samethanks \\
  Meta AI, FAIR, UCL \\
  \texttt{ycxu@meta.com} \\
  \AND
  Bobak Hashemi \\
  Meta AI\\
  \texttt{bobakh@meta.com} \\
  \And
  Lucas Lehnert \\
  Meta AI, FAIR\\
  \texttt{lucaslehnert@meta.com} \\
  \And
  Urun Dogan \\
  Meta AI\\
      \texttt{urundogan@meta.com} \\
  \And
  Zheqing Zhu \\
  Meta AI\\
  \texttt{billzhu@meta.com} \\
  \And
  Olivier Delalleau \\
  Meta AI, FAIR\\
  \texttt{olivier.delalleau@gmail.com} \\
}

\usepackage{wrapfig}

\usepackage{color}
\usepackage{amssymb}
\usepackage{fancyvrb}
\usepackage{pmboxdraw}
\usepackage{amsmath}
\usepackage{amsthm}
\usepackage{tikz}
\usepackage[hang,flushmargin]{footmisc}
\usepackage{chngcntr}
\usepackage[makeroom]{cancel}
\usetikzlibrary{arrows,decorations.markings}

\usepackage{booktabs}
\usepackage{graphicx}
\usepackage{mathtools}
\usepackage{bbm}
\usepackage{relsize}

\let\oldnl\nl
\newcommand{\nonl}{\renewcommand{\nl}{\let\nl\oldnl}}
\makeatletter
\newcommand{\removelatexerror}{\let\@latex@error\@gobble}
\makeatother

\usepackage{enumitem}

\newenvironment{tightlist}%
{\begin{list}{$\bullet$}{%
    \setlength{\topsep}{0in}
    \setlength{\partopsep}{0in}
    \setlength{\itemsep}{0in}
    \setlength{\parsep}{0in}
    \setlength{\leftmargin}{1.5em}
    \setlength{\rightmargin}{0in}
}
}%
{\end{list}
}

\newcommand{\secref}[1]{Section \ref{#1}}

\newcommand{\eqnref}[1]{Eq.~\ref{#1}}
\newcommand{\figref}[1]{Fig.~\ref{#1}}

\newcommand{\tabref}[1]{Table~\ref{#1}}

\newcommand{\appref}[1]{Appendix~\ref{#1}}

\definecolor{green}{RGB}{0, 150, 0}

\newcommand*\samethanks[1][\value{footnote}]{\footnotemark[#1]}

\counterwithin*{thm}{section}

\usepackage{mdframed}
\newmdtheoremenv[nobreak=true]{def2}{Definition}
\renewcommand{\cite}{\citep}

\def\thickhline{%
  \noalign{\ifnum0=`}\fi\hrule \@height \thickarrayrulewidth \futurelet
  \reserved@a\@xthickhline}
\def\@xthickhline{\ifx\reserved@a\thickhline
              \vskip\doublerulesep
              \vskip-\thickarrayrulewidth
             \fi
      \ifnum0=`{\fi}}

\newlength{\thickarrayrulewidth}
\DeclareUnicodeCharacter{225C}{$\triangleq$}
\DeclareUnicodeCharacter{2200}{$\forall$}

\newcommand{\Adv}{A_{\theta}}

\newcommand{\f}{f_{\theta}}

\newcommand{\fM}{f^M_{\theta}}

\newcommand{\h}{h_{\theta}}

\newcommand{\htarg}{h_{\theta^-}}

\newcommand{\ind}{\mathbbm{1}}
\newcommand{\p}{{\pi_{\theta}}}
\newcommand{\pM}{{\pi^M_{\theta}}}

\newcommand{\Q}{Q_{\theta}}
\newcommand{\QM}{Q^M_{\theta}}
\newcommand{\Qtarg}{Q_{\theta^-}}
\newcommand{\rew}{R_{\theta}}
\newcommand{\rewM}{R^M_{\theta}}
\newcommand{\trans}{P}
\newcommand{\V}{V_{\theta}}

\newcommand{\Z}{\R^d}

\renewcommand{\S}{\mathcal{S}} 
\newcommand{\A}{\mathcal{A}}
\newcommand{\R}{\mathbb{R}}

\newcommand{\Loss}{\mathcal{L}}

\newcommand{\E}{\mathbb{E}}

\newcommand{\N}{\mathcal{N}}
\newcommand{\D}{\mathcal{D}}

\DeclarePairedDelimiterX{\infdivx}[2]{(}{)}{%
  #1\;\delimsize\|\;#2%
}

\begin{document}

\maketitle

\begin{abstract}
Model-based reinforcement learning (RL) has shown great promise due to its sample efficiency, but still struggles with long-horizon sparse-reward tasks, especially in offline settings where the agent learns from a fixed dataset. We hypothesize that model-based RL agents struggle in these environments due to a lack of long-term planning capabilities, and that planning in a temporally abstract model of the environment can alleviate this issue. In this paper, we make two key contributions: 1) we introduce an offline model-based RL algorithm, IQL-TD-MPC, that extends the state-of-the-art Temporal Difference Learning for Model Predictive Control (TD-MPC) with Implicit Q-Learning (IQL); 2) we propose to use IQL-TD-MPC as a Manager in a hierarchical setting with \textit{any} off-the-shelf offline RL algorithm as a Worker. More specifically, we pre-train a temporally abstract IQL-TD-MPC Manager to predict ``intent embeddings'', which roughly correspond to subgoals, via planning. We empirically show that augmenting state representations with intent embeddings generated by an IQL-TD-MPC manager significantly improves off-the-shelf offline RL agents' performance on some of the most challenging D4RL benchmark tasks. For instance, the offline RL algorithms AWAC, TD3-BC, DT, and CQL all get zero or near-zero normalized evaluation scores on the medium and large antmaze tasks, while our modification gives an average score over 40.
\end{abstract}

\section{Introduction}
\label{sec:intro}
Model-based reinforcement learning (RL), in which the agent learns a predictive model of the environment and uses it to plan and/or train policies~\cite{worldmodels, planet, schrittwieser2020muzero}, has shown great promise due to its sample efficiency compared to its model-free counterpart~\cite{ye2021efficientzero,micheli2022transformerwm}.
Most prior work focuses on learning single-step models of the world, with which planning can be computationally expensive and model prediction errors may compound over long horizons~\cite{argenson2021modelbased, clavera2020modelaugmented}. As a result, model-based RL still struggles with long-horizon sparse-reward tasks, whereas some evidence suggests that humans are able to combine spatial and temporal abstractions to plan efficiently over long horizons~\cite{botvinick2014mbhrl}.
Modeling the world at a higher level of abstraction can enable predicting long-term future outcomes more accurately and efficiently.

The challenge of long-horizon sparse-reward tasks is particularly prominent in offline RL, where an agent must learn from a fixed dataset rather than from exploring an environment~\cite{levine2020offline,prudencio2023survey,Lange2012BatchRL,ernst2005tree}. The offline setting is key to training RL agents safely, but poses unique challenges such as value mis-estimation~\cite{levine2020offline}.

In this paper, we study offline model-based RL, and hypothesize that planning in a learned temporally abstract model of the environment can produce significant improvements over ``flat'' algorithms that do not use temporal abstraction. Our paper makes two key contributions:
\begin{tightlist}
    \item \textbf{\secref{sec:iqltdmpc}}: We propose IQL-TD-MPC, an offline model-based RL algorithm that combines the state-of-the-art online RL algorithm Temporal Difference Learning for Model Predictive Control (TD-MPC)~\cite{hansen2022tdmpc} with the popular offline RL algorithm Implicit Q-Learning (IQL)~\cite{kostrikov2022iql}. This combination requires several non-trivial design decisions.
    \item \textbf{\secref{sec:hierarchy}}: We show how to use IQL-TD-MPC as a Manager in a temporally abstr                         acted hierarchical setting with \textit{any} off-the-shelf offline RL algorithm as a Worker. To achieve this hierarchy, we pre-train an IQL-TD-MPC Manager to output ``intent embeddings'' via MPC planning, then during Worker training and evaluation, simply concatenate these embeddings to the environment states. These intent embeddings roughly correspond to subgoals\footnote{We generally do not call the intent embeddings ``subgoals'' in this paper because the Worker is not explicitly optimized to achieve them; instead, we are simply concatenating them to environment states.} set $k$ steps ahead, thanks to the coarser timescale used when training the Manager.
    A benefit of this concatenation strategy is its simplicity: it does not require modifying Worker training algorithms or losses.
\end{tightlist}

\begin{figure}[t]
  \centering
    \noindent
    \includegraphics[width=0.8\columnwidth]{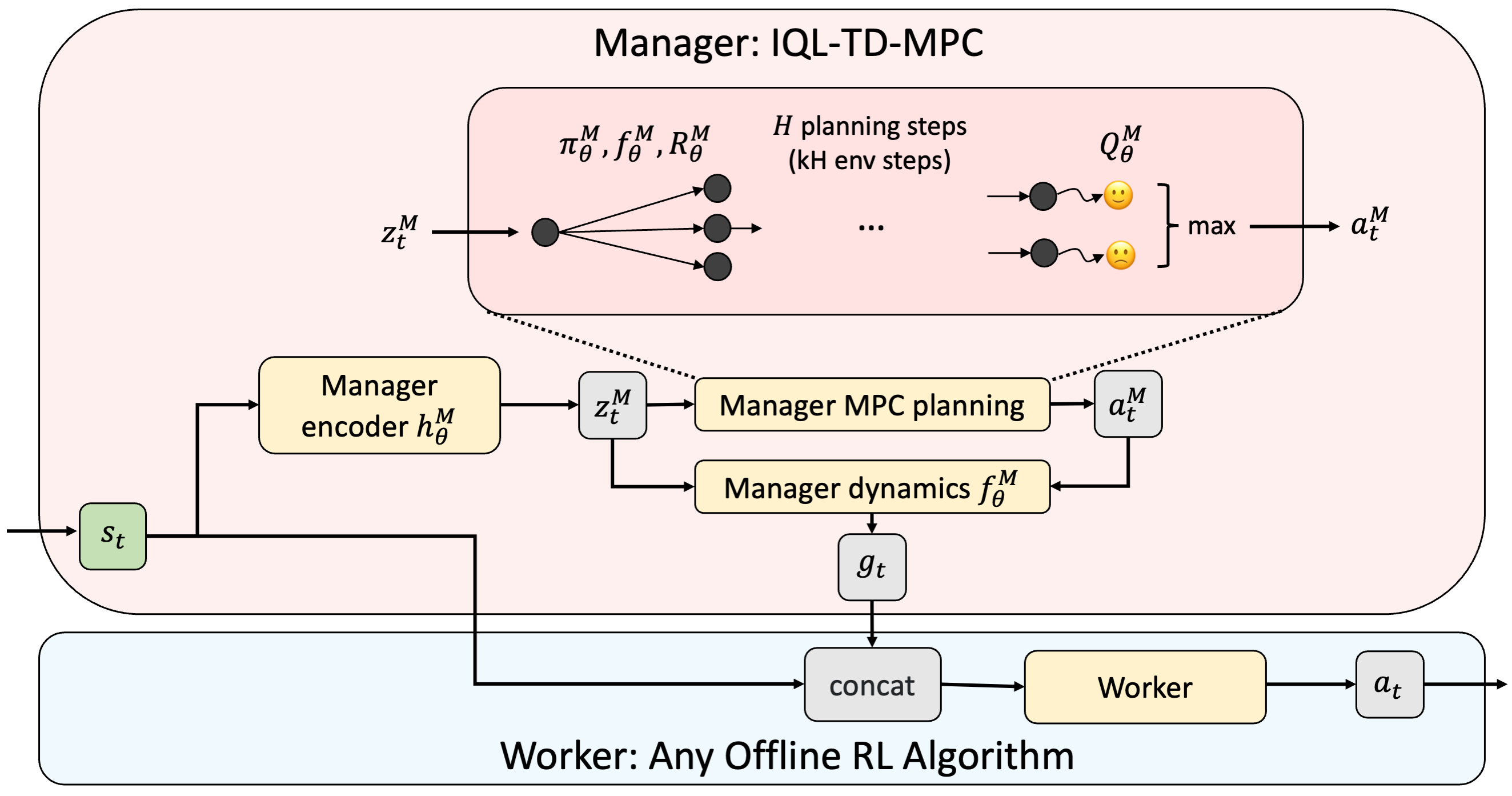}
    \caption{Overview of our hierarchical framework. The Manager is a model-based IQL-TD-MPC agent (inspired by~\citet{kostrikov2022iql} and~\citet{hansen2022tdmpc}) that operates on a coarse timescale to generate intent embeddings $g_t$. To do so, the Manager performs Model Predictive Control over $H$ planning steps (which is $kH$ environment steps), using a learned policy $\pM$, dynamics model $\fM$, reward function $\rewM$, and critic $\QM$. Each intent $g_t$ is concatenated with the state $s_t$ and given to the Worker to output actions $a_t$. This Worker can be any offline RL algorithm.}
  \label{fig:teaser}
\end{figure}

See \figref{fig:teaser} for an overview of our framework. Experimentally, we study the popular D4RL benchmark~\cite{fu2020d4rl}. We begin by showing that IQL-TD-MPC is far superior to vanilla TD-MPC and is on par with several other popular offline RL algorithms. Then, we show the significant benefits of our proposed hierarchical framework. For instance, the well-established offline RL algorithms AWAC~\cite{nair2020awac}, TD3-BC~\cite{fujimoto2021td3bc}, DT~\cite{chen2021dt}, and CQL~\cite{kumar2020cql} all get zero or near-zero normalized evaluation score on the medium and large antmaze variants of D4RL, whereas they obtain an average score of over 40 when used as Workers in our hierarchical framework.
Despite the superior performance of our approach on the maze navigation tasks, our empirical analysis shows that such hierarchical reasoning can be harmful in fine-grained locomotion tasks like the D4RL half-cheetah. 
Overall, our results suggest that model-based planning in a temporal abstraction of the environment can be a general-purpose solution to boost the performance of many different offline RL algorithms, on complex tasks that benefit from higher-level reasoning.
Video results are available at \url{https://sites.google.com/view/iql-td-mpc}.


\section{Preliminaries}
\label{sec:preliminaries}
In this section, we briefly recap the offline RL setting, then provide a detailed review of TD-MPC.

\subsection{Markov Decision Processes and Offline Reinforcement Learning}
\label{sec:mdp}

We consider the standard infinite-horizon Markov Decision Process (MDP)~\cite{puterman1990markov} setting with continuous states and actions, defined by a tuple
$(\S, \A, \trans, R, \gamma, p_0)$ where $\S \subseteq \R^n$ is the state space,
$\A \subseteq \R^m$ is the action space, $\trans(s' \mid s, a)$ is the transition probability distribution function, $R:\mathcal{S} \times \mathcal{A} \mapsto \mathbb{R}$ is the reward function, $\gamma \in (0, 1)$ is the discount factor, and $p_0(s)$ is the initial state distribution function.
The reinforcement learning (RL) objective is to find a policy $\pi(a \mid s)$ that maximizes the expected infinite sum of discounted rewards:
$
\E_{s_0 \sim p_0, a_t \sim \pi(\cdot \mid s_t), s_{t+1} \sim \trans(\cdot \mid s_t, a_t)}\left[\sum_{t=0}^{\infty} \gamma^t R(s_t, a_t)\right]
$.

In offline RL~\cite{levine2020offline,prudencio2023survey}, the agent learns from a fixed dataset rather than collecting its own data in the environment.
One key challenge is dealing with out-of-distribution actions: if the learned policy samples actions for a given state that were not seen in the training set, the model may mis-estimate the value of these actions, leading to poor behavior.
Imitation learning methods like Behavioral Cloning (BC) sidestep this issue by mimicking the behavior
policy used to generate the dataset, but may perform sub-optimally with non-expert data~\cite{hussein2017imitation}.

\subsection{Temporal Difference Model Predictive Control (TD-MPC)}
\label{sec:tdmpc}

Our work builds on TD-MPC~\cite{hansen2022tdmpc}, an algorithm that combines planning in a latent space
using Model Predictive Control (MPC) with actor-critic Temporal Difference (TD) learning.
The components of TD-MPC (with $\theta$ denoting the set of all parameters) are the following:
\begin{tightlist}
\item An encoder $\h: \S \rightarrow \Z$ mapping a state $s$ to its latent representation $z = \h(s)$.
\item A forward dynamics model $\f: \Z \times \A \rightarrow \Z$, predicting the next latent $\hat{z}' = \f(z, a)$.
\item A reward predictor $\rew: \Z \times \A \rightarrow \R$ computing expected rewards $\hat{r} = \rew(z, a)$.
\item A policy $\p: \Z \times \A \rightarrow \R^+$ used to sample actions $a \sim \p(\cdot \mid z)$.
\item A critic $\Q: \Z \times \A \rightarrow \R$ computing state-action values $\Q(z, a)$ that estimate Q-values under $\p$:
$\Q(\h(s), a) \simeq Q^\p(s, a) \triangleq \E_{\p}[\sum_{t \geq 0} \gamma^t \rew(s_t, a_t) \mid s_0 = s, a_0 = a)]$.
\end{tightlist}

The parameters $\theta$ of these components are learned by minimizing several losses over sub-trajectories $(s_0, a_0, r_1, s_1, a_1, \ldots, r_T, s_T)$
sampled from the replay buffer, where $T$ is the horizon:
\begin{tightlist}
\item A critic loss based on the TD error,
$\Loss_Q = ( \Q(\hat{z_t}, a_t) - [r_{t+1} + \gamma \Qtarg(z_{t+1}, \p(z_{t+1}))] )^2$,
where we denote by $\p(z)$ a sample from $\p(\cdot \mid z)$ and $\hat{z}_t = \f(\hat{z}_{t-1}, a_{t-1})$, with $\hat{z}_0 = z_0 = \h(s_0)$.
\item A reward prediction loss, $\Loss_R = (\rew(\hat{z}_t, a_t) - r_{t+1})^2$.
\item A forward dynamics loss (also called ``latent state consistency loss''),
$\Loss_f = \| \f(\hat{z}_t, a_t) - \htarg(s_{t+1}) \|^2$,
where $\theta^-$ are ``target'' parameters obtained by an exponential moving average of $\theta$.
\item A policy improvement loss, $\Loss_\pi = -\Q(\hat{z}_t, \p(\hat{z}_t))$,
only optimized over the parameters of $\p$.
\end{tightlist}

The first three losses are combined through a weighted sum, $\Loss = c_f \Loss_f + c_R \Loss_R + c_Q \Loss_Q$, which trains $\h$, $\f$, $\rew$, and $\Q$. The policy $\p$ is trained independently by minimizing $\Loss_\pi$ without propagating gradients through either $\h$ or $\Q$.



TD-MPC is online: it alternates training the model by minimizing the losses above,
and collecting new data in the environment. At inference time, TD-MPC plans in the latent space with Model Predictive Control, which proceeds in three steps: (1) From current state $s_0$, set the first latent $z_0 = \h(s_0)$, then generate $n_\pi$ action sequences by unrolling the policy $\p$ through the forward model $\f$ over $T$ steps: $a_t \sim \p(\cdot \mid z_t)$ and $z_{t+1} = \f(z_t, a_t)$. (2) Find optimal action sequences using Model Predictive Path Integral (MPPI)~\cite{williams2015mppi}, which iteratively refines the mean and standard deviation of
a Gaussian with diagonal covariance, starting from the above $n_\pi$ action sequences combined with $n_r$ additional random sequences sampled from the current Gaussian.
The quality of an action sequence is obtained by $\sum_{t = 0}^{T-1} \gamma^t \rew(z_t, a_t) + \gamma^T \Q(z_T, a_T)$, i.e., unrolling the forward dynamics model for $T$ steps, using the reward predictor to estimate the sum of rewards, and then bootstrapping with the value estimate of the critic.
(3) One of the best $n_e$ action sequences sampled in the last iteration of the previous step is randomly selected,
and its first action is taken in the environment.

\section{Offline Model-Based RL via Implicit Q-Learning (IQL) and TD-MPC}
\label{sec:iqltdmpc}

In this section, we present IQL-TD-MPC, a framework that extends TD-MPC to the offline RL setting via Implicit Q-Learning (IQL)~\cite{kostrikov2022iql}.
As we show in our experiments, naively training a vanilla TD-MPC agent on offline data performs poorly,
as the model may suffer from out-of-distribution generalization errors
when the training set has limited coverage.
For instance, the state-action value function $\Q$ may ``hallucinate'' very good actions
never seen in the training set, and then the policy $\p$ would learn to predict these actions. This could
steer MPC planning into areas of the latent representation very far from the training distribution, compounding the error further.

IQL addresses the challenge of out-of-distribution actions by combining two ideas:
\begin{tightlist}
    \item Approximating the optimal value functions $Q^*$ and $V^*$ with TD-learning using only actions
    from the training set $\D$.
    This is achieved using the following loss on $\V$:\footnote{For convenience, when describing the original IQL algorithm, we re-use notations $\V$, $\Q$, and $\p$ even though they take raw states $s$ as input rather than latent states $z$.}
    \begin{equation}
    \label{eq:loss_v_iql}
    \Loss_{V,IQL} = \E_{(s, a) \sim \D} [L_2^\tau ( \Qtarg(s, a) - \V(s) )],
    \end{equation}
    where $L_2^\tau$ is the asymmetric squared loss $L_2^\tau(u) = |\tau - \ind_{u < 0}| u^2$,
    and $\tau \in (0.5, 1)$ is a hyper-parameter controlling the ``optimality'' of the learned
    value functions.
    The state-action value function $\Q$ is optimized through the standard one-step TD loss:
    \begin{equation}
    \label{eq:loss_q_iql}
    \Loss_{Q,IQL} = \E_{(s, a, r, s') \sim \D} [(\Q(s, a) - (r + \gamma \V(s')))^2].        
    \end{equation}
    \item Learning a policy using Advantage Weighted Regression~\cite{peng2019awr}, which
    minimizes a weighted behavioral cloning loss whose weights scale exponentially with the advantage:
    \begin{equation}
    \label{eq:loss_pi_iql}
    \Loss_{\pi,IQL} = -\E_{(s, a) \sim \D} [\mathrm{stop\_gradient}(\exp(\beta \Adv(s, a))) \log \p(a \mid s)],
    \end{equation}
    with advantage $\Adv(s, a) = \Qtarg(s, a) - \V(s)$,
    and $\beta >0$ an inverse temperature hyper-parameter.
\end{tightlist}

IQL avoids the
out-of-distribution actions problem by restricting the policy to mimic actions from the data, while still outperforming the behavior policy
by upweighting the best actions under $\Q$ and $\V$.

To integrate IQL into TD-MPC (refer to \secref{sec:tdmpc}), we first replace the TD-MPC policy improvement loss $\Loss_\pi$
with the IQL policy loss $\Loss_{\pi,IQL}$ (\eqnref{eq:loss_pi_iql}). This necessitates training an additional component not present in TD-MPC: a state value function $\V$ to optimize $\Loss_{V,IQL}$ (\eqnref{eq:loss_v_iql}) and $\Loss_{Q,IQL}$ (\eqnref{eq:loss_q_iql}).
As in TD-MPC (and contrary to IQL), all models are applied on learned latent states $z$.
The state-action value function $\Q$ may be trained with either the TD-MPC critic loss $\Loss_Q$
or the IQL critic loss $\Loss_{Q,IQL}$; the difference is whether bootstrapping is done using $\Q$ itself or using $\V$. Our experiments typically use $\Loss_Q$ as we found it to give better results in practice.

This is not enough, however, to fully solve the out-of-distribution actions problem. Indeed, the MPC planning may still prefer actions that
lead to high-return states under $\rew$ and $\Q$, but MPC is actually
exploiting these models' blind spots and ends up performing poorly. We propose the following fix: skip the iterative MPPI refinement of actions during planning, instead keeping only the best $n_e$ sequences of actions among the $n_\pi$ policy samples. This is a special case of the TD-MPC planning algorithm discussed in \secref{sec:tdmpc} where the number of random action sequences $n_r$ is set to zero.

However, this fix brings in another issue: in the original implementation of TD-MPC, actions are sampled
from the policy $\p$ by $a \sim \N(\mu_\theta(z), \sigma^2)$, where $\mu_\theta$ is a learned mean
and $\sigma$ decays linearly towards a fixed hyper-parameter value.
If $\sigma$ is too low, then the policy is effectively deterministic and all $n_\pi$ samples will be nearly identical, which is problematic in our case because we are using $n_r=0$. If $\sigma$ is too high, then
we again run into the problem of out-of-distribution actions.
To avoid having to carefully tune $\sigma$, we learn a stochastic policy that outputs both $\mu_\theta(z)$ and a state-dependent $\sigma_\theta(z)$.\footnote{Our policy implementation is based on the Soft Actor-Critic (SAC) code from~\citet{yarats2020sac}.}

With the above changes (using IQL losses, using only samples from the policy for planning, and
learning a stochastic policy), IQL-TD-MPC preserves TD-MPC's ability to plan efficiently in a learned latent space, while benefiting from IQL's robustness to distribution shift in the offline setting.


\section{IQL-TD-MPC as a Hierarchical Planner}
\label{sec:hierarchy}
We now turn to our second contribution, which is a hierarchical framework (\figref{fig:teaser}) that uses IQL-TD-MPC as a Manager with \textit{any} off-the-shelf offline RL algorithm as a Worker. This hierarchy aims to endow the agent with the ability to reason at longer time horizons.
Indeed, although TD-MPC uses MPC planning to select actions, its planning
horizon is typically short: \citet{hansen2022tdmpc} use a horizon of 5, and found no benefit from increasing it further due to compounding model errors.
For sparse-reward tasks, this makes the planner highly dependent on the quality of the bootstrap estimates predicted by the critic $\Q$, which may be challenging to get right under complex dynamics.

We address this challenge by making IQL-TD-MPC operate as a Manager at a coarser timescale (adding the superscript $M$), processing trajectories $(s_0, a^M_0, r^M_k, s_k, a^M_k, \ldots, r^M_{kH}, s_{kH})$ where:
\begin{tightlist}
    \item $k$ is a hyper-parameter controlling the coarseness of the latent timescale, such that each latent transition skips over $k$ low-level environment steps.
    \item $H$ is the planning horizon; therefore, the effective environment-level horizon is $kH$.
    \item $r^M_{tk} = \sum_{i=(t-1)k+1}^{tk} r_i$, that is, Manager rewards sum up over the previous $k$ environment steps.
    \item $a^M_{tk}$ is an abstract action ``summarizing'' the transition from $s_{tk}$ to $s_{(t+1)k}$.
\end{tightlist}

How should these abstract actions be defined~\cite{pertsch2021accelerating,rosete2023latent}?
Prior work learned an autoencoder that can reconstruct the next latent state~\cite{mandlekar2020iris,li2022gcorl}, and one could define the abstract action as the latent representation of such an autoencoder.
We adopt a similar approach in spirit, but tailored to our TD-MPC setup. Specifically, we train an ``inverse dynamics'' model $b^M_\theta$ in the latent space (instead of the raw environment state space):
\begin{equation}
\label{eq:manageractions}
a^M_{tk} = b^M_\theta(z^M_{tk}, z^M_{(t+1)k}),
\end{equation}
where $z^M_i = h^M_\theta(s_i)$ is the Manager encoding of state $s_i$.
$b^M_\theta$ is trained implicitly by backpropagating through $a_t$ the gradient of the total loss.
Similar to Director~\citep{hafner2022director}, we found discrete actions to help, and thus modify the policy $\pM$ to output discrete actions (see \appref{app:discretemanager} for details).

Once trained, the IQL-TD-MPC Manager can be used to generate ``intent embeddings'' to augment the state representation of any Worker that acts in the environment. We define the intent embedding $g_t \in \Z$ at time $t$ as the difference between the predicted next latent state and the current latent state: 
\begin{equation}
\label{eq:intent_embed}
    g_t = f^M_\theta(z^M_t, a^M_t) - z^M_t, 
\end{equation}
where when training the Worker, $a^M_t$ comes from the inverse dynamics model: $a^M_{t} = b^M_\theta(z^M_{t}, z^M_{t+k})$. Note that we apply the intent embedding on \emph{each} environment step, so \eqnref{eq:intent_embed} does \emph{not} index by $k$.

The Worker can be any policy $\pi$. Its states are concatenated with the intent embeddings: $a_t \sim \pi(\cdot \mid \textsc{concat}(s_t, g_t))$.
Since intent embeddings are in the Manager's latent space, the Manager may be trained independently from the Worker. In practice, we pre-train a single Manager for a task and use it with a range of different Workers (see \secref{sec:exp_universal_manager} for experiments). A benefit of this concatenation strategy is its simplicity: it does not require modifying Worker training algorithms or losses, only appending intent embeddings to states during (i) offline dataset loading and (ii) evaluation.

\subsection{Why are intent embeddings beneficial for offline RL?}
\label{subsec:intuitive}

Before turning to experiments, we provide an intuitive explanation for why we believe augmenting states with intent embeddings can be beneficial. For simplicity, we focus on the well-understood Behavioral Cloning (BC) algorithm, but we note that many other offline RL algorithms such as Advantage Weighted Actor-Critic (AWAC)~\cite{nair2020awac}, Implicit Q-Learning (IQL)~\cite{kostrikov2022iql}, and Twin Delayed DDPG Behavioral Cloning (TD3-BC)~\cite{fujimoto2021td3bc} use the BC objective in some way, and thus the intuition may carry over to these algorithms as well.


We first provide an information-theoretic argument to explain why intent embedding should make the imitation learning objective easier to optimize. One of the primary obstacles in long-horizon sparse-reward offline RL is the ambiguity surrounding the relationship between each state-action pair in a dataset and its corresponding long-term objective. By incorporating intent embeddings derived from MPC planning into state-action pairs, our framework provides offline RL algorithms with a more well-defined association between each state-action pair and the objective being targeted.
For a BC policy $\pi: \mathcal{S} \mapsto \mathcal{A}$, we typically train $\pi$ to match the state-action pairs in the offline dataset. With intent embeddings, the agent can instead learn $\pi': \mathcal{S} \times \Z \mapsto \mathcal{A}$, which maps a pair of state and intent random variables $(S_t, G_t)$ to an action random variable $A_t$. Since $A_t$ is not independent of $G_t$ given $S_t$, the mutual information $I((S_t, G_t); A_t) \geq I(S_t; A_t)$, so $(S_t, G_t)$ contains at least as much information about $A_t$ as $S_t$ does on its own when learning a BC policy via imitation learning.

The above argument explains why it should be easier to optimize the BC objective when training the Worker, thanks to the additional information contained in the intent embedding. This can be particularly beneficial on offline datasets built from a mixture of varied policies~\cite{fu2020d4rl}.
In addition to simplifying the task of the BC Worker, the Manager is trained to provide ``good'' intent embeddings at inference time. This is achieved through the MPC-based planning procedure of IQL-TD-MPC, by identifying a sequence of abstract actions $(a^M_t, a^M_{t+k}, \ldots, a^M_{t+kH})$ that leads to high expected return (according to $\rewM$ and $\QM$ when unrolling $\fM$).
The intent embedding $g_t$, obtained from $a^M_t$ through \eqnref{eq:intent_embed}, is then used to condition the Worker policy, similar to prior work on goal-conditioned imitation learning~\citep{mandlekar2020iris,lynch2020playlmp}.


\section{Experiments}
\label{sec:experiments}
Our experiments aim to answer three questions: \textbf{(Q1)} How does IQL-TD-MPC perform as an offline RL algorithm, compared to both the original TD-MPC algorithm and other offline RL algorithms? \textbf{(Q2)} How much benefit do we obtain by using IQL-TD-MPC as a Manager in a hierarchical setting? \textbf{(Q3)} To what extent are the observed benefits actually coming from our IQL-TD-MPC algorithm?

\textbf{Experimental Setup.} We focus on continuous control tasks of the D4RL benchmark~\cite{fu2020d4rl},
following the experimental protocol from CORL~\cite{tarasov2022corl}: training with a batch size of 256 and reporting the  normalized score (0 is random, 100 is expert) at end of training, averaged over 100 evaluation episodes.
Averages and standard deviations are reported over 5 random seeds. Each experiment was run on an A100 GPU. Training for a single seed (including both pre-training the Manager and training the Worker) took $\sim 5$ hours on average.
See \appref{app:hyperparameters} for all hyper-parameters.

\subsection{(Q1) How does IQL-TD-MPC perform as an offline RL algorithm?}

\begin{table*}[t]
\resizebox{\columnwidth}{!}{
\begin{tabular}{c|c|c|c|c|c}
\hline
\diagbox{Dataset}{Algorithm}& IQL        &TT               & TAP                                   & TD-MPC            & IQL-TD-MPC \\
\hline\hline
antmaze-umaze-v2            & $87.5\pm2.6$  & $\textbf{100.0}\pm0.0$   & $81.5\pm2.8$           & $44.6\pm28.2$         & $52.0\pm46.0$          \\
antmaze-umaze-diverse-v2    & $\textbf{66.2}\pm13.8$    & $21.5\pm2.9$    & $\textbf{68.5}\pm3.3$     & $0.0\pm0.0$         & $\textbf{72.6}\pm26.6$ \\
antmaze-medium-play-v2      & $71.5\pm12.6$    & $\textbf{93.3}\pm6.4$         & $78.0\pm4.4$    & $1.8\pm3.91$     & $\textbf{88.8}\pm5.9$ \\
antmaze-medium-diverse-v2   & $70.0\pm10.9$   & $\textbf{100.0}\pm0.0$          & $85.0\pm3.6$          & $0.0\pm0.0$ & $40.3\pm34.2$   \\
antmaze-large-play-v2       & $40.8\pm12.7$      & $66.7\pm12.2$       & $\textbf{74.0}\pm4.4$          & $0.0\pm0.0$ & $66.6\pm13.7$     \\
antmaze-large-diverse-v2    & $47.5\pm9.5$      & $60.0\pm12.7$        & $\textbf{82.0}\pm5.0$           & $0.0\pm0.0$ & $4.0\pm4.1$       \\
antmaze-ultra-play-v0       & $9.2\pm6.7$       & $\textbf{20.0}\pm10.0$              &  $\textbf{22.0}\pm4.1$           & $0.0\pm0.0$      & $\textbf{20.6}\pm16.0$         \\
antmaze-ultra-diverse-v0    & $\textbf{22.5}\pm8.3$       & $\textbf{33.3}\pm12.2$             &  $\textbf{26.0}\pm4.4$           & $0.0\pm0.0$      & $3.6\pm10.1$          \\
\hline
maze2d-umaze-v1             & $37.7\pm2.0$              & $36.7\pm2.1$      & $\textbf{58.6}\pm1.4$     & $\textbf{76.4}\pm20.8$          & $40.9\pm45.3$ \\
maze2d-medium-v1            & $35.5\pm1.0$              & $32.7\pm1.1$        & $-3.9\pm0.3$         & $85.3\pm15.8$            & $\textbf{161.0}\pm11.3$ \\
maze2d-large-v1             & $49.6\pm22.0$             & $33.2\pm1.0$           & $-2.1\pm0.1$       & $\textbf{121.6}\pm27.0$          & $\textbf{158.9}\pm77.1$ \\
\hline
halfcheetah-medium-v2       & $48.3\pm0.11$    & $46.9\pm0.4$         & $45.0\pm0.1$             & $45.7\pm14.6$      & $\textbf{57.4}\pm0.1$  \\
halfcheetah-medium-replay-v2& $44.2\pm1.2$       & $41.9\pm2.5$       & $40.8\pm0.6$            & $45.7\pm5.0$      & $\textbf{49.2}\pm1.3$  \\
halfcheetah-medium-expert-v2 & $94.6\pm0.2$  & $\textbf{95.0}\pm0.2$   & $91.8\pm0.8$     & $-1.0\pm0.9$     & $44.8\pm8.5$           \\
\hline
\end{tabular}}
\caption{Normalized scores of IQL-TD-MPC, other offline RL algorithms (IQL~\cite{kostrikov2022iql}, TT~\cite{janner2021tt}, TAP~\cite{jiang2022tap}) and TD-MPC on D4RL after 1M training steps. IQL results are from \citet{tarasov2022corl}. TT and TAP results are from their papers, except for antmaze-umaze-diverse and maze2d, which we reproduced with the default hyperparameters because they were not reported. Each entry shows the mean over 100 episodes and 5 seeds, and the standard deviation over seeds.
Bolded numbers are within one standard deviation of the best result in each row.
}
\label{tab:d4rl_all}
\end{table*}

We begin with a preliminary experiment to verify that IQL-TD-MPC is a viable offline RL algorithm. For this experiment, we compare IQL-TD-MPC, TD-MPC, and several offline RL algorithms from the literature on various tasks. See \tabref{tab:d4rl_all} for results. There are several key trends we can observe. First, vanilla TD-MPC does not perform well in general, and completely fails in the more difficult variants of the antmaze task. This is expected because TD-MPC is not designed to train from offline data. The one exception is the umaze environment in maze2d, where TD-MPC actually outperforms IQL-TD-MPC by a significant margin. We hypothesize that this is because the dynamics of this environment are very simple, and the data provides adequate coverage to learn effective TD-MPC models, while the conservative expectile updates of IQL-TD-MPC cause learning to be slower.
The other trend we see is that IQL-TD-MPC is generally on par with the other offline RL algorithms. This confirms our hypothesis that IQL-TD-MPC is a viable model-based offline RL algorithm.

\subsection{(Q2) How much benefit do we obtain by using IQL-TD-MPC as a Manager?}
\label{sec:exp_universal_manager}

\begin{table*}[t]
\centering
\resizebox{\columnwidth}{!}{
\begin{tabular}{c|c|c|c|c|c|c}
\hline
\diagbox{Dataset}{Algorithm}& AWAC        & BC            & DT            & IQL           & TD3-BC & CQL    \\
\hline\hline
antmaze-umaze-v2    & {\color{green}$51 \to 86$} & {\color{green}$52 \to 78$}  & {\color{green}$64 \to 89$} & {\color{green}$44 \to 80$} & {\color{red}$90 \to 82$} & $67 \to 69$  \\
antmaze-umaze-diverse-v2    & $53 \to 60$ & $49 \to 48$  & $55 \to 38$ & $60 \to 51$ & $45 \to 53$ & $37 \to 36$  \\
antmaze-medium-play-v2    & {\color{green}$0 \to 36$} & {\color{green}$0 \to 52$}  & {\color{green}$0 \to 43$} & $70 \to 64$ & {\color{green}$0.2 \to 60$} & {\color{green}$0.8 \to 33$}  \\
antmaze-medium-diverse-v2    & {\color{green}$0.8 \to 16$} & {\color{green}$0.2 \to 20$}  & {\color{green}$0.2 \to 33$} & {\color{red}$63 \to 30$} & {\color{green}$0.4 \to 21$} & {\color{green}$0.2 \to 14$}  \\
antmaze-large-play-v2       & {\color{green}$0 \to 67$} & {\color{green}$0 \to 50$}  & {\color{green}$0 \to 53$} & $54 \to 70$ & {\color{green}$0 \to 46$} & {\color{green}$0 \to 19$} \\
antmaze-large-diverse-v2    & {\color{green}$0 \to 40$} & {\color{green}$0 \to 38$}  & {\color{green}$0 \to 31$} & $ 31 \to 46$ & {\color{green}$0 \to 29$}  & {\color{green}$0 \to 16$}  \\
antmaze-ultra-play-v0       & {\color{green}$0 \to 18$} & {\color{green}$0 \to 18$}  & {\color{green}$0 \to 10$} & $9 \to 16$ & {\color{green}$0 \to 20$} & {\color{green}$0 \to 5$} \\
antmaze-ultra-diverse-v0    & {\color{green}$0 \to 37$} & {\color{green}$0 \to 35$}  & {\color{green}$0 \to 10$} & $22 \to 27$ & {\color{green}$0 \to 29$} & {\color{green}$0.6 \to 5$}  \\
\hline
maze2d-umaze-v1    & $77 \to 78$ & {\color{green}$3 \to 64$}  & {\color{green}$26 \to 63$} & {\color{green}$41 \to 77$} & {\color{green}$39 \to 77$} & $-14 \to 7$  \\
maze2d-medium-v1       & $43 \to 67$ & {\color{green}$3 \to 70$}  & {\color{green}$13 \to 71$} & {\color{green}$32 \to 78$} & $101 \to 47$ & {\color{red}$104 \to 16$} \\
maze2d-large-v1    & {\color{red}$193 \to 132$} & {\color{green}$-1 \to 94$} & {\color{green}$3 \to 96$} & {\color{green}$42 \to 135$} & $69 \to 126$ & $53 \to 64$ \\
\hline
halfcheetah-medium-v2    & {\color{red}$49 \to 45$} & {\color{green}$42 \to 45$}  & {\color{green}$42 \to 47$} & {\color{red}$47 \to 43$} & {\color{red}$47 \to 44$} & {\color{red}$46 \to 44$} \\
halfcheetah-medium-replay-v2       & {\color{red}$45 \to 41$} & {\color{green}$34 \to 40$}  & $39 \to 37$ & {\color{red}$44 \to 40$} & {\color{red}$44 \to 39$} & {\color{red}$45 \to 32$} \\
halfcheetah-medium-expert-v2    & {\color{red}$95 \to 80$} & {\color{green}$57 \to 84$}  & $63 \to 52$ & {\color{red}$92 \to 79$} & $86 \to 76$  & {\color{red}$90 \to 45$} \\
\hline
\end{tabular}}
\caption{Results of our hierarchical framework, where we append IQL-TD-MPC Manager intents to states in various offline RL algorithms taken from the CORL repository~\cite{tarasov2022corl}.
Each table entry is of the form ``baseline evaluation score $\to$ our evaluation score''. In either case, we report scores after 500K steps of training; in general, we found that all agents plateaued after this point. For our hierarchical framework, these 500K steps correspond to 300K steps of pre-training the Manager, followed by 200K steps of training the CORL Worker. All entries report a mean over 5 independent random seeds; see \tabref{tab:corl_main_results_std} in \appref{app:stds} for standard deviations. Green entries indicate statistically significant improvement, while red entries indicate statistically significant degradation.
}
\label{tab:corl_main_results}
\end{table*}

Now, we turn to the main results of our work, where we demonstrate the benefits of using IQL-TD-MPC as a Manager with a range of different non-hierarchical offline RL algorithms as Workers. For this experiment, we used offline RL algorithms from the CORL repository~\cite{tarasov2022corl} as Workers.
We concatenated intent embeddings output by the Manager to the environment states seen by these Workers during both training and evaluation. Once the boilerplate code was written, the
changes to the CORL algorithms were straightforward, since
they typically only required adding two lines of code to (i) augment states in the offline dataset and (ii) wrap the evaluation environment.

\tabref{tab:corl_main_results} shows the results of this experiment, for the following CORL Workers: Advantage Weighted Actor-Critic (AWAC)~\cite{nair2020awac}, Behavioral Cloning (BC), Decision Transformer (DT)~\cite{chen2021dt}, Implicit Q-Learning (IQL)~\cite{kostrikov2022iql}, Twin Delayed DDPG Behavioral Cloning (TD3-BC)~\cite{fujimoto2021td3bc}, and Conservative Q-Learning (CQL)~\cite{kumar2020cql}.
Overall, we observe a dramatic improvement in performance for all these agents compared to their baseline versions, whose only difference is the lack of intent embeddings concatenated to state vectors. Interestingly, vanilla AWAC / BC / DT / TD3-BC all get a zero score on the large and ultra variants of the antmaze task, while with our modification, they are able to learn to solve the task.
This shows that the intent embeddings produced by the Manager are highly useful, and can be used to compensate for the lack of long-term planning abilities in off-the-shelf RL agents.


Notably, our approach slightly worsens performance on the half-cheetah locomotion tasks. A likely explanation is that these tasks are more about fine-grained control and thus have less natural hierarchical structure for our framework to exploit. The intent embeddings are trained by having the Manager look at states $k$ timesteps ahead, but lookahead may not help on these tasks.
We hypothesize that they may actually hurt as they restrict the pool of candidate actions the Worker is considering.

\begin{figure}[t]
  \centering
    \noindent
    \includegraphics[width=0.9\columnwidth]{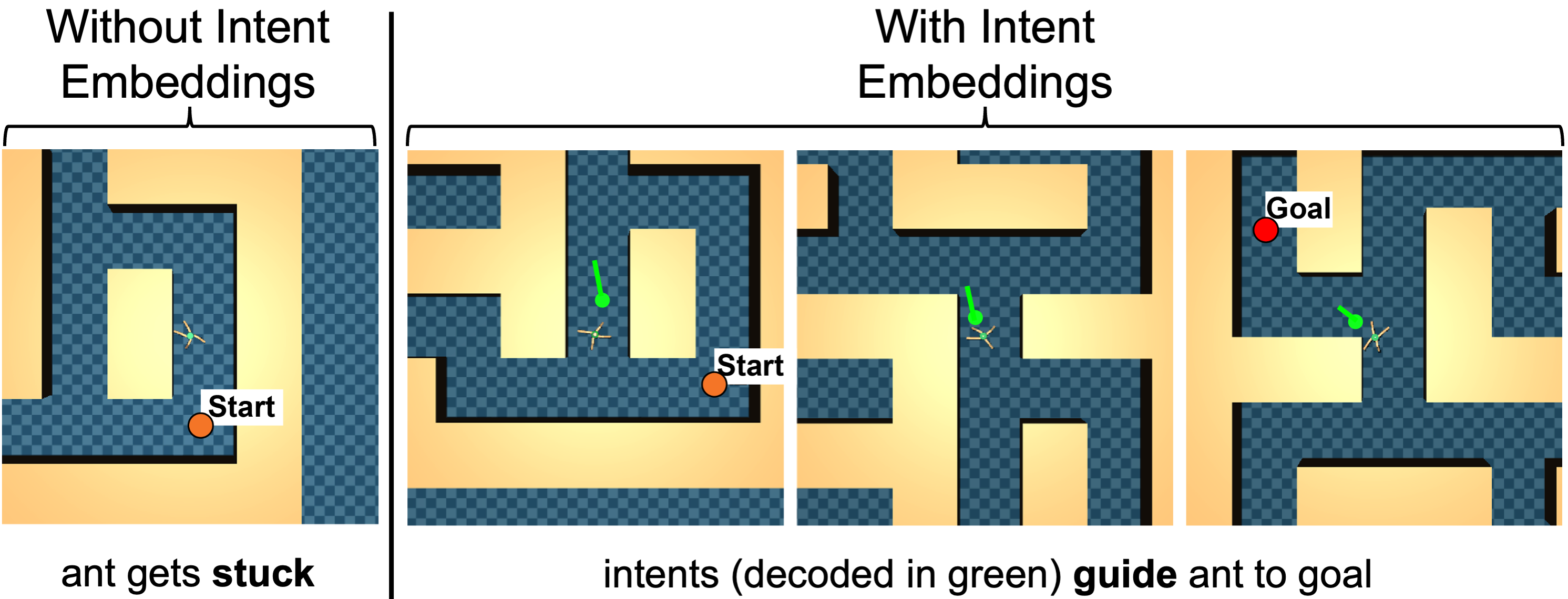}
    \caption{Visualization of an episode of the Behavioral Cloning (BC) agent on the antmaze-large-play-v2 task. On the left, without intent embeddings, the ant gets stuck close to the start of the maze, never reaching the goal. On the right, the ant reaches the goal, guided by the intent embeddings whose decoding is visualized in green. We see that the intent embeddings act as latent-space subgoals.}
  \label{fig:episode}
\end{figure}

In \figref{fig:episode}, we visualize an episode of the Behavioral Cloning (BC) agent on the antmaze-large-play-v2 task, in order to qualitatively understand the benefits of our framework. On the left, we see that without intent embeddings, the ant gets stuck close to the start of the maze, never reaching the goal. On the right, we see that the ant reaches the goal, guided by the intent embeddings visualized in green. To generate these green visualizations, we trained a separate decoder alongside the IQL-TD-MPC Manager that converts the intent embeddings (in the Manager's latent space) back into the raw environment state space, which contains the position and velocity of the ant. The green dot shows the position, and the green line attached to the dot shows the velocity (speed is the length of the line). This decoder was trained on a reconstruction loss and did not affect the training of the other models. Overall, this visualization shows that the intent embeddings are effectively acting as latent-space subgoals that the Worker exploits to learn a more effective policy.

\subsection{(Q3) To what extent are the observed benefits coming from IQL-TD-MPC?}

\begin{table*}[t]
\centering
\resizebox{0.85\columnwidth}{!}{
\begin{tabular}{c|c|c|c|c}
\hline
\diagbox{Dataset}{Algorithm}& AWAC        & BC                     & IQL           & TD3-BC   \\
\hline\hline
antmaze-medium-play-v2    & $0 \to 0$ & $0 \to 0$  & $70 \to 66$ & $0.2 \to 0$  \\
antmaze-medium-diverse-v2    & $0.8 \to 0.2$ & $0.2 \to 0$  & $63 \to 71$ & $0.4 \to 0.2$  \\
antmaze-large-play-v2       &  $0 \to 0$ & $0 \to 0$  & {\color{red}$54 \to 25$} & $0 \to 0$  \\
antmaze-large-diverse-v2    &  $0 \to 0$ & $0 \to 0$  & $ 31 \to 37$ & $0 \to 0$  \\
\hline
halfcheetah-medium-v2    & $49 \to 49$ & $42 \to 42$  & $47 \to 47$ & $47 \to 47$  \\
halfcheetah-medium-replay-v2       & $45 \to 43$ & $34 \to 34$  & $44 \to 43$ & $44 \to 44$  \\
halfcheetah-medium-expert-v2    & $95 \to 93$ & $57 \to 61$  & $92 \to 90$ & $86 \to 87$  \\
\hline
\end{tabular}}
\caption{Ablation results, where we replace Manager intent embeddings with random vectors. Each table entry is of the form ``baseline evaluation score $\to$ ablation evaluation score''. We report scores after 500K steps of training. All entries report a mean over 5 independent random seeds; see \tabref{tab:random_vector_ablation_std} in \appref{app:stds} for standard deviations. Red entries indicate statistically significant degradation.
}
\label{tab:random_vector_ablation}
\end{table*}

One may wonder whether the strong results in \tabref{tab:corl_main_results} are simply due to a ``regularization'' effect, or whether the intent embeddings simply tie-break the stochasticity of the behavior policy. We conduct an ablation to address this: we run our framework, but replace the intent embeddings with random vectors of the same dimensionality, with entries drawn uniformly from $(0, 1)$. See \tabref{tab:random_vector_ablation} for results.

Across nearly all tasks and algorithms, we found no statistically significant difference between the baseline and the ablation. This means that the Workers typically learned to ignore the random vectors. Comparing against the clear benefits of our proposed method in \tabref{tab:corl_main_results}, we can conclude that IQL-TD-MPC was critical; it guides the Workers in a more impactful way than just regularization.

Interestingly, \tabref{tab:random_vector_ablation} shows that the Workers learned to ignore the random vectors in the half-cheetah tasks (performance is unchanged), while in \tabref{tab:corl_main_results}, our modification \textit{harmed} performance. This confirms that the intent embeddings are correlated with environment states in a way that RL algorithms do not ignore, which may help or hurt depending on how much hierarchical structure the task has.

\section{Related Work}
\label{sec:related}



\subsection{Offline Reinforcement Learning}

In offline reinforcement learning~\cite{levine2020offline,prudencio2023survey,Lange2012BatchRL,ernst2005tree}, the agent learns from a fixed offline dataset.
\citet{li2022gcorl} learn a generative model of potential goals to pursue given the current state, along with a goal-conditioned policy trained by Conservative Q-Learning (CQL,~\citealt{kumar2020cql}), from a combination of the task reward with a goal-reaching reward.
Planning is performed by optimizing goals (with CEM) to maximize those rewards as estimated by the value function of the policy over the planning horizon.
In our work, by contrast, our intent embeddings are defined in the Manager's learned latent space, and we can use this Manager with any offline RL Worker.
The recently proposed POR algorithm~\cite{xu2022por} learns separate ``guide'' and ``execute'' policies, where the ``guide'' policy abstracts out the action space.
Our Manager can also be seen as such a guide that would plan over longer time horizons.



A recent line of work uses Transformers~\cite{vaswani2017attention} to model the trajectories in the offline dataset~\cite{janner2021tt,chen2021dt}.
\citet{jiang2022tap} propose the Trajectory Autoencoding Planner (TAP), that models a trajectory by a sequence of discrete tokens learned by a Vector Quantised-Variational AutoEncoder (VQ-VAE, \citealp{vandenoord2017vqvae}), conditioned on the initial state.
This enables efficient search with a Transformer-based trajectory generative model.
One can relate this approach to ours by interpreting the generation of encoded trajectories as the Manager, and the decoding into actual actions as the Worker.
However, this distinction is somewhat artificial since in contrast to our approach, the Manager provides an intent embedding that encodes an entire predicted trajectory, rather than a single state.
In addition, TAP relies on a Monte-Carlo ``return-to-go'' estimator to bootstrap search, while we explicitly learn a temporally abstract Manager value function.


Play-LMP~\cite{lynch2020playlmp} encodes goal-conditioned sub-trajectories in a latent space through a conditional sequence-to-sequence VAE~\cite{sohn2015structuredcvae}, which can be used to sample latent plans that are decoded through a goal-conditioned policy.
However, there is no notion of optimizing a task reward here: instead, the desired goal state must be provided as input to the model to solve a task.

\subsection{Hierarchical Reinforcement Learning}

Though our work focuses on the offline setting, we highlight a few related works in the online setting.
Director~\cite{hafner2022director} trains Manager and Worker policies in imagination, where the Manager actions are discrete representations of goals for the Worker, learned from the latent representation of a world model.
Although we re-use a similar discrete representation for Manager actions, this approach differs from our work in several ways: it focuses on the online setting, there is no planning during inference, and the world model is not temporally abstract.
Our work may be related to the literature on option discovery~\cite{sutton1999between,bagaria2020option,daniel2016probabilistic,brunskill2014pac}. In our proposed hierarchical framework, the intent embeddings output by our Manager can be seen as latent skills~\cite{pertsch2021accelerating,rosete2023latent} that the Worker conditions on to improve its learning efficiency.
Finally, our work can be seen as an instantiation of one piece of the H-JEPA framework laid out by \citet{lecun2022path}: we learn a Manager world model at a higher level of temporal abstraction, which works in tandem with a Worker to optimize rewards. 


\section{Limitations and Future Work}
\label{sec:conclusion}

In this paper, we propose a non-trivial extension of TD-MPC to the offline setting based on IQL, and leverage its superior planning abilities as a temporally extended Manager in a hierarchical architecture.
Our experiments confirm the benefits of this hierarchical framework in guiding offline RL algorithms.

Our algorithm still suffers from a number of limitations that we intend to tackle in future work: (1) Our method hurts performance on some locomotion tasks (\tabref{tab:corl_main_results}), which require fine-grained control. It is unsurprising that hierarchy does not help in such contexts; however, further investigation is required to confirm our intuition for why the Worker algorithms are unable to simply ignore these harmful intent embeddings. (2) The Worker agent may also be improved by actively planning toward the intent embedding set by the Manager. For instance, the Worker itself could be an IQL-TD-MPC agent modeling the world at the original environment timescale, unlike the temporally abstract Manager. (3) Our Manager's timescale is defined by a fixed hyper-parameter $k$. This could instead be set dynamically by the Manager, and included in the intent embedding concatenated to the environment state. (4) Similar to the TD-MPC algorithm we build on, our approach is computationally intensive, both during Manager pre-training and inference, because we need to unroll the Manager's world model to obtain the intent embeddings. A potential avenue to speed it up could be to represent the world model as a Transformer~\cite{vaswani2017attention,micheli2022transformerwm}, for more efficient rollouts.

\bibliography{main}
\bibliographystyle{plainnat}


\newpage
\appendix

\section*{Appendix}

\section{Discrete Manager Actions in IQL-TD-MPC}
\label{app:discretemanager}
Similar to \citet{hafner2022director}, we define the Manager's action as a vector of several categorical
variables. The inverse dynamics model $b^M_\theta$ (\eqnref{eq:manageractions}) takes latent states $z^M_{tk}$ and $z^M_{(t+1)k}$ as input and outputs a matrix of $L \times C$ logits, representing $L$ categorical distributions with $C$ categories each. The model then samples a $C$-dimensional one-hot vector from each of the $L$ distributions, and flattens the results into a sparse binary vector of length $L \times C$. See Figure G.1 from \citet{hafner2022director} for a visualization. This sparse binary vector serves as the ``action'' chosen by the Manager. The model is optimized end-to-end together with all other components of IQL-TD-MPC using straight-through gradients~\cite{bengio2013estimating}. Unlike \citet{hafner2022director}, we did not include a KL-divergence regularization term in the model objective as we found regularizing the distribution towards some uniform prior hurts the final performance. In all our experiments, we used $L=8$ and $C=10$.

As a result of this change, we must also change the Manager policy network $\pM$ to output discrete actions. Hence, we modify $\pM$ to output $L$ categorical distributions of size $C$ (applying a softmax instead of the squashed Normal distribution used for continuous actions in IQL-TD-MPC). The  behavioral cloning term $\log \p(a \mid s)$ in the IQL policy loss (\eqnref{eq:loss_pi_iql}) thus becomes a cross-entropy loss over the $C$ categories. This loss is summed over the $L$ categorical distributions, which are treated independently.

\section{Standard Deviation Tables}
\label{app:stds}

We provide standard deviations accompanying the results in the main text. \tabref{tab:corl_main_results_std} provides standard deviations for \tabref{tab:corl_main_results}, and \tabref{tab:random_vector_ablation_std} provides standard deviations for \tabref{tab:random_vector_ablation}.

\begin{table*}[h]
\centering
\resizebox{\columnwidth}{!}{
\begin{tabular}{c|c|c|c|c|c|c}
\hline
\diagbox{Dataset}{Algorithm}& AWAC        & BC            & DT            & IQL           & TD3-BC & CQL   \\
\hline\hline
antmaze-umaze-v2    & $9 \to 12$ & $6 \to 2$  & $4 \to 3$ & $4 \to 4$ & $3 \to 3$ & $7 \to 3$  \\
antmaze-umaze-diverse-v2    & $10 \to 8$ & $5 \to 7$  & $6 \to 11$ & $7 \to 18$ & $6 \to 3$ & $21 \to 3$ \\
antmaze-medium-play-v2    & $0 \to 13$ & $0 \to 8$  & $0 \to 11$ & $5 \to 6$ & $0.4 \to 4$  & $0.8 \to 10$  \\
antmaze-medium-diverse-v2    & $1 \to 9$ & $0.4 \to 8$  & $0.5 \to 11$ & $6 \to 7$ & $0.5 \to 1$ & $0.4 \to 6$ \\
antmaze-large-play-v2       & $0 \to 10$ & $0 \to 3$  & $0 \to 3$ & $9 \to 7$ & $0 \to 8$ & $0 \to 5$  \\
antmaze-large-diverse-v2    & $0 \to 5$ & $0 \to 5$  & $0 \to 7$ & $11 \to 9$ & $0 \to 1$ & $0 \to 5$   \\
antmaze-ultra-play-v0       & $0 \to 8$ & $0 \to 4$  & $0 \to 2$ & $6 \to 8$ & $0 \to 6$ & $0 \to 2$ \\
antmaze-ultra-diverse-v0    & $0 \to 12$ & $0 \to 8$ & $0 \to 3$ & $8 \to 18$ & $0 \to 7$  & $0.8 \to 2$  \\
\hline
maze2d-umaze-v1    & $38 \to 5$ & $4 \to 9$  & $12 \to 1$ & $1 \to 3$ & $14 \to 2$ & $0.8 \to 42$  \\
maze2d-medium-v1       & $21 \to 21$ & $5 \to 7$  & $3 \to 7$ & $7 \to 10$ & $49 \to 8$ & $15 \to 42$ \\
maze2d-large-v1    & $20 \to 35$ & $0.5 \to 8$  & $2 \to 8$ & $21 \to 30$ & $21 \to 66$  & $61 \to 90$ \\
\hline
halfcheetah-medium-v2    & $0.2 \to 0.1$ & $0.2 \to 0.2$  & $0.3 \to 0.5$ & $0.3 \to 0.4$ & $0.2 \to 0.3$   & $0.1 \to 0.1$ \\
halfcheetah-medium-replay-v2       & $0.1 \to 0.3$ & $0.7 \to 0.7$  & $0.2 \to 1$ & $0.3 \to 1$ & $0.2 \to 1$ & $0.4 \to 5$ \\
halfcheetah-medium-expert-v2    & $0.8 \to 9$ & $6 \to 4$  & $7 \to 5$ & $0.9 \to 7$ & $8 \to 5$  & $2 \to 2$  \\
\hline
\end{tabular}}
\caption{Standard deviations accompanying the means reported in \tabref{tab:corl_main_results}. The table is formatted in the same way, so all these standard deviations are in the same positions as their corresponding means.
}
\label{tab:corl_main_results_std}
\end{table*}

\begin{table*}[ht]
\centering
\begin{tabular}{c|c|c|c|c}
\hline
\diagbox{Dataset}{Algorithm}& AWAC        & BC                     & IQL           & TD3-BC   \\
\hline\hline
antmaze-medium-play-v2    & $0 \to 0$ & $0 \to 0$  & $5 \to 2$ & $0.4 \to 0$ \\
antmaze-medium-diverse-v2    & $1 \to 0.4$ & $0.4 \to 0$  & $6 \to 7$ & $0.5 \to 0.4$  \\
antmaze-large-play-v2       &  $0 \to 0$ & $0 \to 0$  & $9 \to 6$ & $0 \to 0$  \\
antmaze-large-diverse-v2    & $0 \to 0$ & $0 \to 0$  & $11 \to 12$ & $0 \to 0$  \\
\hline
halfcheetah-medium-v2    & $0.2 \to 0.3$ & $0.2 \to 0.2$  & $0.3 \to 0.1$ & $0.2 \to 0.4$  \\
halfcheetah-medium-replay-v2       & $0.1 \to 0.4$ & $0.7 \to 1$  & $0.3 \to 0.7$ & $0.2 \to 0.4$  \\
halfcheetah-medium-expert-v2    & $0.8 \to 2$ & $6 \to 5$  & $0.9 \to 2$ & $8 \to 5$  \\
\hline
\end{tabular}
\caption{Standard deviations accompanying the means reported in \tabref{tab:random_vector_ablation}. The table is formatted in the same way, so all these standard deviations are in the same positions as their corresponding means.
}
\label{tab:random_vector_ablation_std}
\end{table*}

\section{Hyper-parameters}
\label{app:hyperparameters}

In this section, we list all hyper-parameters used in experiments.
\tabref{tab:hyper_tdmpc} contains hyper-parameters that were already present in the original TD-MPC algorithm (or that we added to slightly tweak its behavior, e.g., the ability to disable Prioritized Experience Replay or to use the policy mean in the TD target instead of a sample).
Changes compared to the original TD-MPC implementation (\url{https://github.com/nicklashansen/tdmpc}) are bolded.

\tabref{tab:hyper_iqltdmpc} lists the hyper-parameters for IQL-TD-MPC, related to integrating the IQL losses and making the continuous policy $\p$ stochastic. \tabref{tab:hyper_hierarchy} lists the hyper-parameters for using IQL-TD-MPC as a Manager (using a discrete stochastic policy $\p$, as discussed in \appref{app:discretemanager}).

\begin{table}[htbp]
\centering
\begin{tabular}{c|c|c}
\hline
\textbf{hyper-parameter}          & \textbf{Value in TD-MPC}  & \textbf{Value in IQL-TD-MPC}  \\
\hline
$\gamma$                        & 0.99              & 0.99\\
latent dimension                & 50                & 50 \\
planning horizon $H$            & \bf{5}            & \bf{2} \\
CEM population size             & 512               & 512\\
CEM \#policy actions ($n_\pi$)  & \textbf{25}       & \textbf{512}\\
CEM \#random actions ($n_r$)    & \textbf{487}        & \textbf{0}\\
CEM elite size ($n_e$)          & 64                & 64\\
CEM iterations                  & 6                 & 6\\
CEM momentum coefficient        & 0.1               & 0.1\\
CEM temperature                 & 0.5               & 0.5\\
enable Prioritized Experience Replay & \textbf{yes}  & \textbf{no}\\
learning rate                   & $\bm{10^{-3}}$ & $\bm{3\cdot10^{-4}}$ \\
batch size                      & \textbf{512}      & \textbf{256}\\
MLP hidden size                 & 512               & 512\\
encoder / decoder hidden size   & 256               & 256\\
bootstrapping value on last planning state& $\bm{\Q(s, \p(s_H))}$ & $\bm{\Q(s, \E_{a \sim \p(a | s_H)}[a]})$ \\
bootstrapping value in TD target& $\bm{\Q(s', \p(s'))}$ & $\bm{\Q(s', \E_{a' \sim \p(a' | s')}[a']})$ \\
reward loss coefficient ($c_R$) & 0.5               & 0.5\\
critic loss coefficient ($c_Q$) & 0.1               & 0.1\\
consistency loss coefficient ($c_f$) & 2            & 2\\
temporal coefficient ($\rho$)    & 0.5              & 0.5\\
gradient clipping threshold     & 10                & 10\\
$\theta^-$ update frequency     & 2                 & 2\\
$\theta^-$ update momentum      & 0.01              & 0.01\\
\hline
\end{tabular}
\caption{TD-MPC hyper-parameters that we use in our IQL-TD-MPC algorithm.
Bolded values are those that were modified compared to the original TD-MPC implementation from \citet{hansen2022tdmpc}.
We found no benefit to increasing the planning horizon beyond 2 in the offline setting.
The motivation for changing $n_\pi$ and $n_r$ is described in \secref{sec:iqltdmpc}.
We disabled Prioritized Experience Replay out of caution in the offline setting, to be sure that the initial arbitrary priority (assigned to all transitions in the buffer after loading the dataset) would not artificially bias the sampling distribution (a problem that does not occur in the online setting, where each new transition gets assigned the maximum priority seen so far).
Decreasing the learning rate and using the policy mean for bootstrapping were found to lead to more stable results for some tasks.
Using a smaller batch size was purely for the purpose of fair comparison with prior results reported in the literature.
}
\label{tab:hyper_tdmpc}
\end{table}

\begin{table}[htbp]
\centering
\begin{tabular}{c|c}
\hline
\textbf{hyper-parameter}    & \textbf{Value in   IQL-TD-MPC} \\
\hline
IQL $\tau$                          & 0.9 \\
IQL $\beta$                         & $3$ \\
exponential advantage threshold & 100 \\
loss for critic $\Q$           & $\Loss_Q$ (exception: $\Loss_{Q,IQL}$ for antmaze-\{medium,large,ultra\}-* tasks)  \\
critic loss $\Loss_{V,IQL}$ coefficient & 0.1 \\
stochastic policy $\log \sigma$ (std) range & $(-5,2 )$ \\
stochastic policy action clipping threshold & 0.99 \\
stochastic policy entropy bonus weight    & 0.1 \\
\hline
\end{tabular}
\caption{Hyper-parameters that we introduced specifically for our IQL-TD-MPC algorithm in the ``flat'' (non-hierarchical) setting described in \secref{sec:iqltdmpc}. These hyper-parameters were used to obtain the results in \tabref{tab:d4rl_all}.
The action clipping threshold clips actions from the offline dataset to avoid infinite loss $\Loss_{\pi,IQL}$ (\eqnref{eq:loss_pi_iql}).
The entropy bonus weight is the coefficient of an extra term we add to $\Loss_{\pi,IQL}$ to maximize entropy so as to prevent policy collapse.
This term is approximated as $\log \p(s)$, where $\p(s)$ is a random action sampled from $\p(\cdot \mid s)$.
}
\label{tab:hyper_iqltdmpc}
\end{table}

\begin{table}[htbp]
\centering
\begin{tabular}{c|c}
\hline
\textbf{hyper-parameter}    & \textbf{Value in IQL-TD-MPC when used as a Manager} \\
\hline
latent dimension        & 10 \\
planning horizon $H$                 & 4 \\
reward scale factor & 0.1 for maze2d and locomotion tasks, 1.0 for antmaze tasks\\
IQL $\tau$                          & 0.9 \\
IQL $\beta$                         & $3 / \text{reward scale factor}$ \\
exponential advantage threshold & 100 \\
loss for critic $\Q$           & $\Loss_Q$ (exception: $\Loss_{Q,IQL}$ for antmaze-\{medium,large,ultra\}-* tasks)  \\
critic loss $\Loss_{V,IQL}$ coefficient & 0.1 \\
latent timescale coarseness $k$                             & 8 \\
discrete policy $L$ (\appref{app:discretemanager}) & 8 \\
discrete policy $C$ (\appref{app:discretemanager}) & 10 \\
\hline
\end{tabular}
\caption{Hyper-parameters that were used specifically for our IQL-TD-MPC algorithm in the setting described in \secref{sec:hierarchy} where IQL-TD-MPC is a Manager. These hyper-parameters were used to obtain the results in \tabref{tab:corl_main_results} and \tabref{tab:random_vector_ablation}.
Compared to ``flat'' IQL-TD-MPC (\tabref{tab:hyper_tdmpc} and \tabref{tab:hyper_iqltdmpc}), we decreased the latent dimension as we found no benefit in using higher values, while increasing the planning horizon for the Manager proved useful.
The reward scale factor scales down manager rewards in tasks where otherwise summing rewards over $k$ timesteps can lead to high Q-values and an explosion of critic losses.
The IQL inverse temperature $\beta$ is also updated accordingly to ``cancel out'' the effect of this rescaling in the advantage weight computation.
}
\label{tab:hyper_hierarchy}
\end{table}

\end{document}